\documentclass[sigconf]{acmart}

\AtBeginDocument{%
  \providecommand\BibTeX{{%
    \normalfont B\kern-0.5em{\scshape i\kern-0.25em b}\kern-0.8em\TeX}}}

\copyrightyear{2021} 
\acmYear{2021} 
\setcopyright{acmcopyright}\acmConference[KDD '21]{Proceedings of the 27th ACM SIGKDD Conference on Knowledge Discovery and Data Mining}{August 14--18, 2021}{Virtual Event, Singapore}
\acmBooktitle{Proceedings of the 27th ACM SIGKDD Conference on Knowledge Discovery and Data Mining (KDD '21), August 14--18, 2021, Virtual Event, Singapore}
\acmPrice{15.00}
\acmDOI{10.1145/3447548.3467390}
\acmISBN{978-1-4503-8332-5/21/08}

\usepackage{mathtools}
\usepackage{multirow}
\usepackage{colortbl}
\usepackage[ruled,vlined]{algorithm2e}
\usepackage{graphicx}
\usepackage{subfigure}
\usepackage{caption}

\newcommand{\vect}[1]{\mathbf{#1}}
\newcommand{\matr}[1]{\mathbf{#1}}
\newcommand{\method}{\texttt{PETGEN}}

\usepackage[english]{babel}
\usepackage{blindtext}

\begin{document}

\title{PETGEN: Personalized Text Generation Attack on \\
Deep Sequence Embedding-based Classification Models}

\author{Bing He, Mustaque Ahamad, Srijan Kumar}
\affiliation{%
  \institution{Georgia Institute of Technology}
  \city{Atlanta}
  \state{Georgia}
  \country{USA}
}
\email{bhe46@gatech.edu,
mustaq@cc.gatech.edu, 
srijan@gatech.edu}

\begin{abstract}
\textit{What should a malicious user write next to fool a detection model?}
Identifying malicious users is critical to ensure the safety and integrity of internet platforms. Several deep learning based detection models have been created. 
However, malicious users can evade deep detection models by manipulating their behavior, rendering these models of little use. The vulnerability of such deep detection models against adversarial attacks is unknown.
Here we create a novel adversarial attack model against deep user sequence embedding-based classification models, which use the sequence of user posts to generate user embeddings and detect malicious users.
In the attack, the adversary generates a new post to fool the classifier. 
We propose a novel end-to-end Personalized Text Generation Attack model, called \texttt{PETGEN}, that simultaneously reduces the efficacy of the detection model and generates posts that have several key desirable properties. Specifically, \texttt{PETGEN} generates posts that are personalized to the user's writing style, have knowledge about a given target context, are aware of the user's historical posts on the target context, and encapsulate the user's recent topical interests. 
We conduct extensive experiments on two real-world datasets (Yelp and Wikipedia, both with ground-truth of malicious users) to show that \texttt{PETGEN} significantly reduces the performance of popular deep user sequence embedding-based classification models. \texttt{PETGEN} outperforms five attack baselines in terms of text quality and attack efficacy in both white-box and black-box classifier settings. 
Overall, this work paves the path towards the next generation of adversary-aware
sequence classification models.
\vspace{-2mm}
\end{abstract}








\fancyhead{} 

\maketitle

\vspace{-3mm}
\section{Introduction}
As Web platforms, such as e-commerce, social media, and crowdsourcing platforms, have gained popularity, they are increasingly targeted by malicious actors for their gains~\cite{noorshams2020ties, kumar2018rev2, kumar2017army}. The proliferation of undesirable users, such as fake accounts~\cite{noorshams2020ties}, spammers~\cite{dou2020robust, rayana2015collective}, fake news spreaders~\cite{shu2017fake, le2020malcom}, abnormal users~\cite{chalapathy2019deep}, vandal editors~\cite{kumar2015vews},  fraudsters~\cite{kumar2018rev2}, and sockpuppets~\cite{kumar2017army}, poses a threat to the safety and integrity of online communities. To give an example, on Facebook, roughly 5\% of monthly active users in 2019 were fake accounts~\cite{noorshams2020ties}. Similarly, on Amazon, 63\% reviews on beauty products were from fraudulent users~\cite{teodora2021}. Thus, the identification of malicious accounts is a critical task for all web and social media platforms. 

\begin{figure}[tbp!]
    \centering
    \includegraphics[width=\columnwidth]{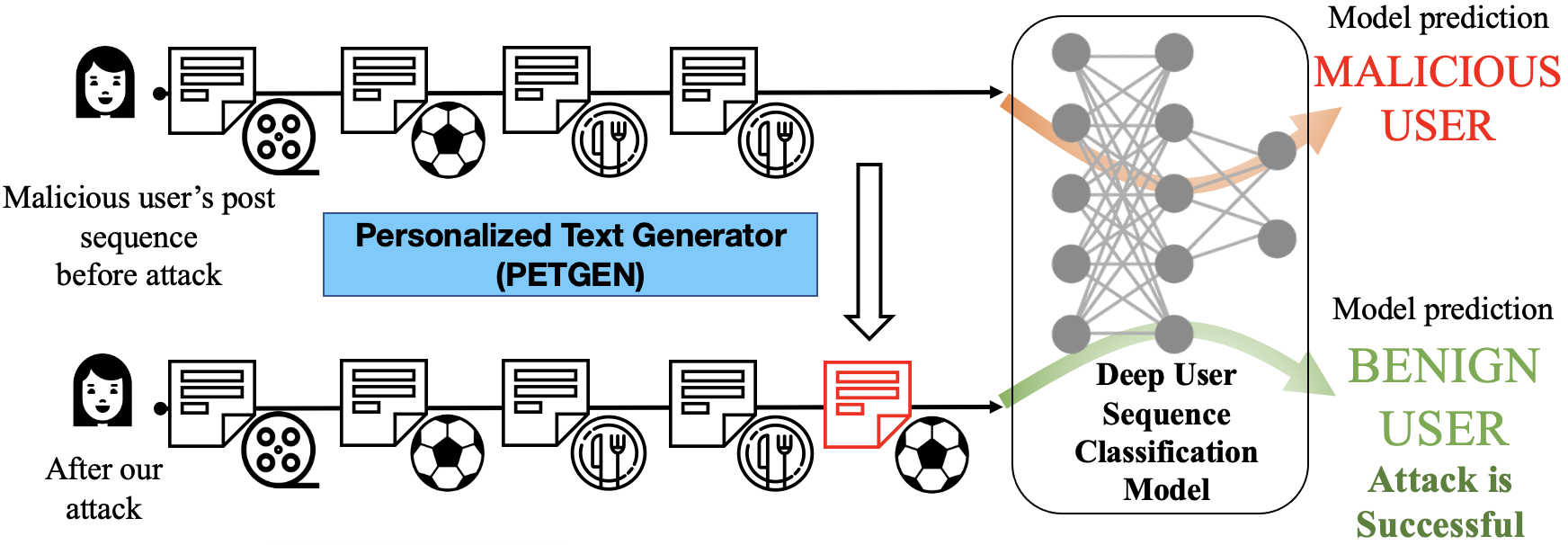}
    \vspace{-20pt}
    \caption{
    Deep user sequence embedding-based classification models are used to detect malicious users (top row). However, an evasion attack by an adversary by creating a new fake post can lead the same model to misclassify it as a benign user (bottom row). Our method, \method{}, generates personalized text posts to adversarially attack the classifier.
    }
    \vspace{-18pt}
    \label{fig:setting}
\end{figure}

Deep user sequence embedding-based classification models are increasingly gaining popularity for platform integrity tasks, including the TIES model at Facebook~\cite{noorshams2020ties}. 
These models train a deep learning model to generate user embeddings by utilizing the temporal sequence of actions and post content of a user. The user embedding is then used to make predictions about the user. For example, Figure~\ref{fig:setting} shows a deep user sequence embedding-based classification model trained to identify malicious users from the user's sequence of posts (top row).

\begin{figure*}[h]
    \centering
    \subfigure[\vspace{-12pt}F1 score after attack] {
        \includegraphics[width=0.15\textwidth]{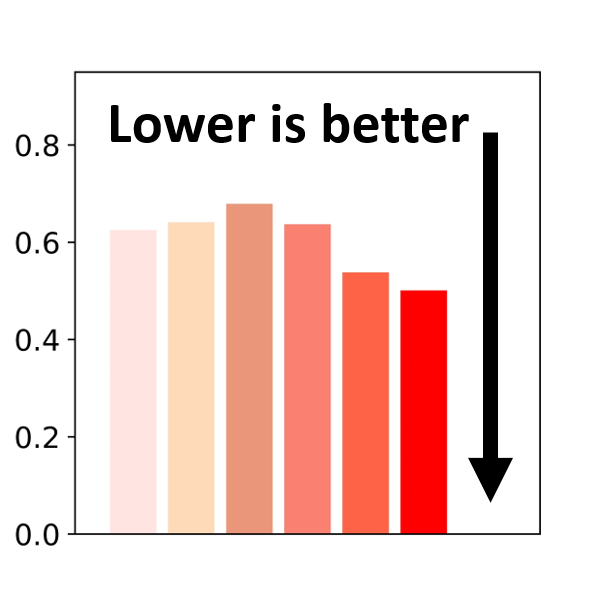}
    }
    \subfigure[\vspace{-12pt}Attack Rate] {
        \includegraphics[width=0.15\textwidth]{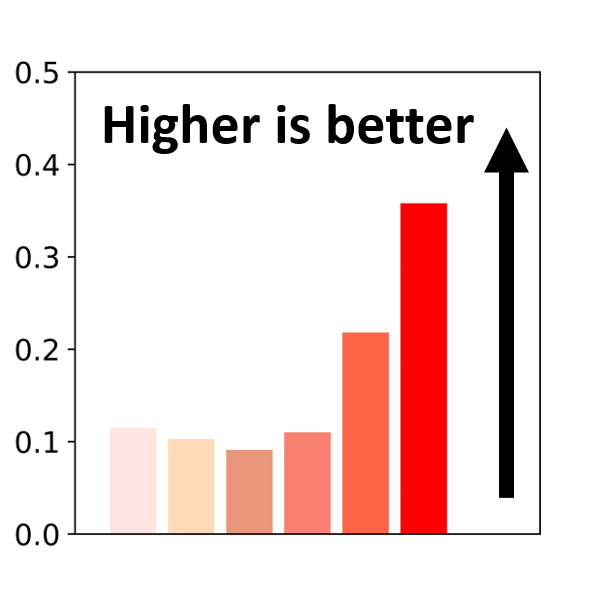}
    }
    \subfigure[\vspace{-12pt}BLEU score] {
        \includegraphics[width=0.15\textwidth]{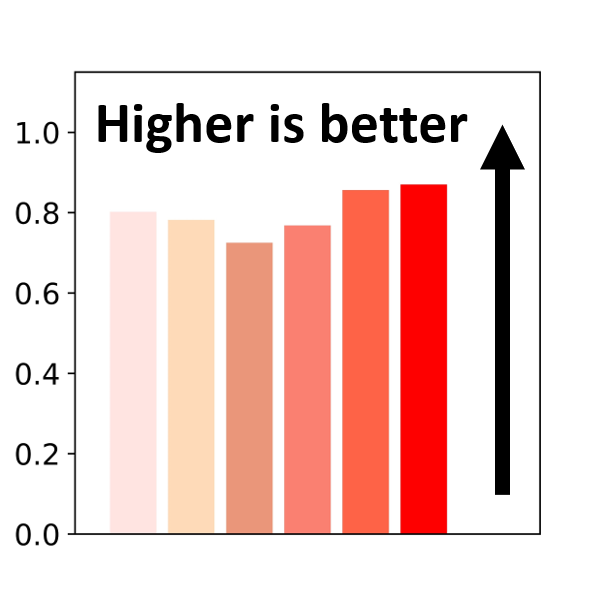}
    }
    \subfigure[\vspace{-12pt}Target Context Similarity] {
        \includegraphics[width=0.15\textwidth]{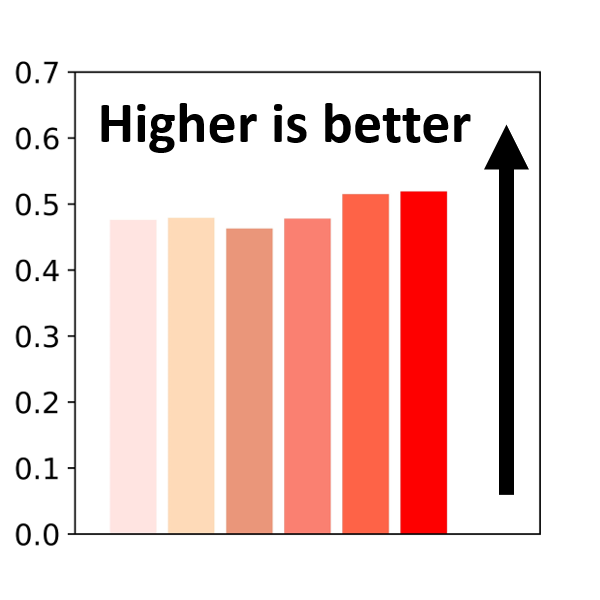}
    }
    \subfigure[\vspace{-12pt}Recent Post Similarity] {
        \includegraphics[width=0.15\textwidth]{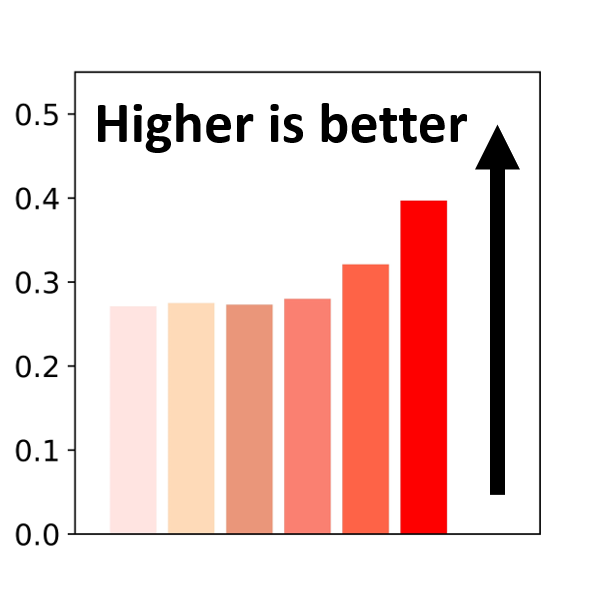}
    }
    \subfigure[\vspace{-12pt}Context Post Similarity] {
        \includegraphics[width=0.15\textwidth]{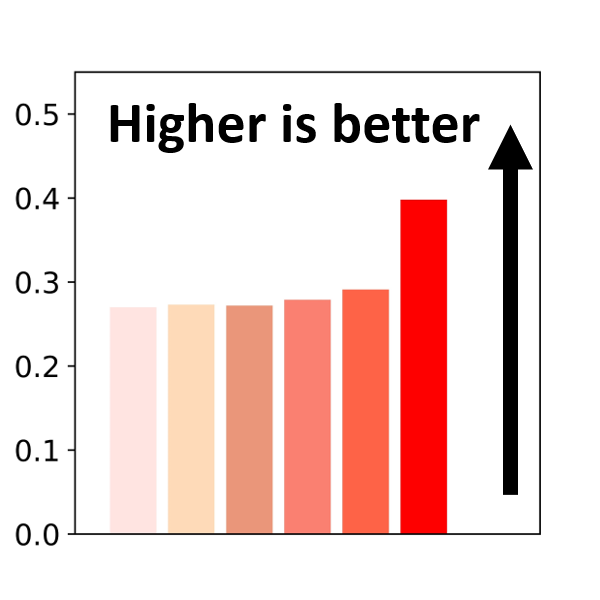}
    }
    \\
    \vspace{-6pt}
    \subfigure{
        \includegraphics[width=0.60\textwidth]{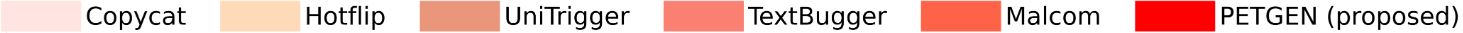}
    }
    \vspace{-11pt}
    \caption{Comparison between the performance  of \method{} and existing attack methods on the fake reviewer detection model in the Yelp dataset. \method{} performs the best by reducing the F1 score after attack to the lowest value and its attack rate is the highest. Simultaneously, \method{} generates better quality text by having the highest BLUE score, and target context, recent post, and context post similarities. 
    }
    \vspace{-12pt}
    \label{fig:showcase_result}
\end{figure*}

However, deep learning models can be vulnerable to adversarial attacks~\cite{papernot2016limitations}. While adversarial attacks on deep learning models have received a lot of attention in graph representation learning, natural language processing, and computer vision domains~\cite{papernot2016limitations}, the vulnerability of deep user sequence embedding-based classification models remains unknown. For example, in Figure~\ref{fig:setting}, the malicious user can create a new post, so that the entire user sequence is misclassified as benign by the classifier (bottom row). 
Thus, identifying the vulnerabilities of deep user sequence classification models is crucial to improve the models for real-world robustness. 

In this paper, we conduct an adversarial evasion attack on deep user sequence embedding-based classification models. Our \textbf{attack setting} is as follows: given a pre-trained deep user sequence classification model $\mathcal{F}$ (trained to classify users as malicious or benign),
a user's sequence of posts, 
and a target topic context, 
the goal of the attacker is to generate a new post on the target context such that the entire user sequence is now misclassified by $\mathcal{F}$.

Generating a fake attack post poses three major \textbf{challenges}. 
First, how can the text generation process effectively use the user's post sequence, such that the generated post aligns with the user's historical posts on similar contexts? 
Second, how to generate adversarial text that can fool a sequence embedding-based classifier? 
Finally, how to generate text that is personalized? Specifically, how can the text capture the user's writing style, be aware of user's recent vs past interests, and be knowledgeable about target context.

\textbf{Existing text generation methods} suffer from three major shortcomings with respect to our attack setting: 
(a) recent work has adversarially attacked fake news detection classifiers by generating fake reply comments on the posts~\cite{le2020malcom}. However, these models can not be directly used to attack sequence classification models, as generated text is not personalized to the user and thus, can be identified by anomaly detection models~\cite{chalapathy2019deep}. 
(b) Adversarial posts generated by flipping characters~\cite{ebrahimi2017hotflip} or minimally changing characters or words~\cite{li2018textbugger} can be easily detected by robust detection systems that employ spell-checkers or human evaluators. Attacks that append short random text or phrases to the original text can also be detected by topic coherence checkers~\cite{le2020malcom}.
(c) Many attacks require editing existing text, which is not always possible in real attack settings (\textit{e.g.}, on Twitter, tweets can not be edited once written). Our attack setting requires creation of a new post altogether.

\textbf{Present work.} In this work, we create a \underline{Pe}rsonalized \underline{T}ext \underline{Gen}eration attack framework, called \method{}, to generate adversarial text to attack deep user sequence embedding-based classification models. \method{} is an end-to-end model. It leverages the sequential history of user posts (solution to challenge 1) by utilizing the relationship between the user's historical posts and the target context, and builds a context-biased user sequence embedding. This is used to generate an initial version of the attack post. Next, the model adopts a multi-stage multi-task learning approach to manipulate the text to effectively attack the classification model (solution to challenge 2) and personalize the text to the user's writing style, recent interests, and make the text relevant to the global discussions in the target topic context (solution to challenge 3). This step outputs the final attack text of \method{}.

We evaluate the attack effectiveness and text quality of our model. 
We use two popular datasets: Yelp fake reviewer dataset~\cite{rayana2015collective} and Wikipedia vandal editor dataset~\cite{kumar2015vews}, both with ground truth malicious users. 
We evaluate two popular deep user sequence embedding-based classification models: TIES, a model that is used in production at Facebook~\cite{noorshams2020ties} and HRNN, a sequence classification model that uses sequential text embedding~\cite{lee2016sequential}. 
We compare \method{} against five baseline and recent attack models that can generate attack text. 
Experiments reveal several key findings. 
First, both deep user sequence classification models are vulnerable to the fake text generation attack. Their model performance drops with even one generated post. 
Second, \method{} generates attack text that results in a larger classification performance drop compared to existing attack methods.
Third, the text generated by \method{} has higher quality and is more personalized than existing attack methods. Experimental results on Yelp dataset are in Figure~\ref{fig:showcase_result}.
Fourth, \method{} is highly effective in both the white-box setting (when the attacker has access to the details of the classification model) and the black-box attack setting (when the attacker does not know anything about the classification model). 
Finally, human evaluators rate text generated by \method{} as being more realistic over text generated by existing generation-based attack methods.

Overall, our main contributions are:
\begin{itemize}
    \item \textbf{New attack setting:} To the best of our knowledge, we are the first to investigate the problem of text generation attack on deep user sequence embedding-based classifiers, where adversaries generate a new piece of text added at the end of post sequence to fool the sequential classifier.
    \item \textbf{Attack model:} We create \method{}, a multi-stage multi-task personalized text generation model that can generate attack that can effectively attack the sequence classifier and generate high-quality personalized text. 
    \item \textbf{Effectiveness:} Extensive experiments on two datasets show that our methods can outperform five strong baselines in terms of the attack performance.
    Moreover, our method generates text with higher quality, both in terms of quantifiable metrics and as evaluated by human evaluators. 
\end{itemize}

The code and data are at: \url{http://claws.cc.gatech.edu/petgen}.

\section{Related Work}

\subsection{Deep User Sequence Classification Models}
To determine whether a user is malicious or not, existing methods usually focus on building deep sequence embedding models to encode the sequential information and use the embedding for downstream applications~\cite{zhao2019recurrent,Kumar2019PredictingDE, lee2016sequential}. For example, Facebook first creates a temporal embedding from users' sequence of posts, then predicts users' dynamic embedding when users write a new post, and finally uses these embeddings for fake account detection~\cite{noorshams2020ties}. 
However, vulnerabilities of these deep user sequence embedding-based classification models have not been explored. To fill this gap, in this work, we first introduce a new next post generation attack on these models and further propose a attack framework to conduct this attack. Our work reveals the vulnerabilities of these models. 

\begin{table}[tbp!]
    \centering
    \begin{tabular}{c|p{0.8\linewidth}}
    \hline
         Notation & Description  \\
         \hline
         $\vect{p}_u^t $    & User $u$'s post at time $t$ \\
         $\matr{P}_u^{1:T}$     & User $u$'s sequence of past $T$ posts \\
         $\hat{\vect{p}}_u^{T+1} $    & User $u$'s generated post at time $T+1$ \\
         $ \vect{c}_u^{t} $ & User $u$'s context for post $\vect{p}_u^t$ \\
         $ \matr{C}_u^{1:T} $ & User $u$'s sequence of contexts $ \vect{c}_u^{t} ,  t\in\{1, ..., T\}$\\
         $\vect{b}_u$ & The target context for user $u$\\
         $y_u$ & The ground truth label of user $u$  \\
         $\mathcal{G}$ & The text generator \\
         $\mathcal{F}$ & The pre-trained user sequence classifier \\
          \hline
    \end{tabular}
    \caption{Table of notations used in the paper.}
    \vspace{-25pt} 
    \label{tab:symbol}
\end{table}
\subsection{Sequential Text Classification Models} 
Many works formulate classification of the sequence of a user's posts as a sequential text classification~\cite{lee2016sequential,  yang2016hierarchical} or document classification problem~\cite{yang2016hierarchical}. 
In practice, Convolutional Neural Network (CNN) and RNN are widely utilized to capture the sequential reliance between text posts and encode the text features for detection~\cite{lee2016sequential}. 
However, these sequential text classification models can be vulnerable to adversarial attacks, which is relatively unexplored.

\subsection{Adversarial Text Generation}

Generating adversarial text to attack text classifiers is an important task due to its contribution to model robustness~\cite{zhang2020adversarial}.
These methods can mainly be grouped into two categories: 
\noindent (1) \textit{Modification-based attacks}: these approaches mainly make minor modifications to existing text to generate new text. Modifications include changing or adding characters, words, or phrases ~\cite{ebrahimi2017hotflip, Wallace2019Triggers, li2018textbugger, jia2019towards}. 

However, these models have various shortcomings: they are incapable of fully leveraging a user's rich history of posts, they can not generate original content, and their modifications can be easily detected by finding misspelled words and improperly manipulated sentences~\cite{pruthi2019combating}. 
\noindent (2) \textit{Generation-based attacks}: these methods (e.g., TextGAN~\cite{nie2018relgan}) generate a new piece of text to achieve the attack goal. 
A recent attack model called Malcom~\cite{le2020malcom} generates new fake reply comments to news articles to fool detectors. This model achieves high success in fooling the detector. However, these attack models have some shortcomings: they are not designed to leverage a user's rich history of posts and the generated text is not personalized to the user. 
To overcome the above drawbacks of both the generation-based and modification-based  methods, we propose \method{}, an end-to-end personalized text generation model that leverages user sequences to output personalized posts to effectively fool classifiers.

\section{Problem Definition}
In this section, we formally define our problem as follows:

\textbf{Preliminaries:} 
We are given $N$ users ${U} = \{u_1, ... u_N\}$ and a set of user ground truth labels $\mathcal{Y} = \{y_u\}$, 
where $y_u=0$ means user $u$ is a benign user and $y_u=1$ means $u$ is a malicious user.
For each user $u$, we are given a sequence of chronologically ordered posts $\matr{P}_u^{1:T}=\{\vect{p}_u^1,...,\vect{p}_u^t,..., \vect{p}_u^T \}, \matr{P}_u^{1:T} \in \mathcal{R}^{T \times d} $ where $\vect{p}_u^t \in \mathcal{R}^d $ denotes user $u$'s post at time $t$ and $d$ is the number of tokens in the post.
Each post has an associated context, describing the topic, background, or metadata of the post in detail. So, the sequence of contexts is $\matr{C}_u^{1:T} =\{ \vect{c}_u^1,..., \vect{c}_u^t, ..., \vect{c}_u^T  \}, \matr{C}_u^{1:T} \in \mathcal{R}^{T \times d'} $ where $\vect{c}_u^t$ is the topic context of post $\vect{p}_u^t$ and $d'$ is the number of tokens in context. 
We are given a pre-trained deep user sequence embedding-based classification model $\mathcal{F}$, which generates user $u$'s predicated label $\mathcal{F}(\matr{P}_u^{1:T})$. Model $\mathcal{F}$ is trained to predict $\mathcal{F}(\matr{P}_u^{1:T}) = y_u, \forall u \in {U}$.

\textbf{Attacker goal:}
Given user $u$'s sequence of posts $\matr{P}_u^{1:T}$, contexts $\matr{C}_u^{1:T}$, ground truth label $y_u$, and target context $\vect{b}_u$, we aim to generate next post $\hat{\vect{p}}_u^{T+1}$,
such that
$\mathcal{F}([{\matr{P}}_u^{1:T}, \hat{\vect{p}}_u^{T+1})]) = 1-{y}_u$. Here $[{\matr{P}}_u^{1:T}, \hat{\vect{p}}_u^{T+1}]$ represents a sequence where the post $\hat{\vect{p}}_u^{T+1}$ is concatenated at the end of the sequence ${\matr{P}}_u^{1:T}$.
Thus, the goal of the attacker is to flip the prediction result of the classifier on the user's original post sequence. 
Our modeling goal is to train a text generator $\mathcal{G}$ that generates the post $\hat{\vect{p}}_u^{T+1}$ using the user's historical posts. 
Thus, $\hat{\vect{p}}_u^{T+1} = \mathcal{G}(\matr{P}_u^{1:T}, \matr{C}_u^{1:T}, \vect{b}_u)$.  We list the symbols in Table~\ref{tab:symbol}.

\section{Methodology}
\subsection{System Overview}
\begin{figure}[tbp]
    \centering
    \includegraphics[width=\linewidth]{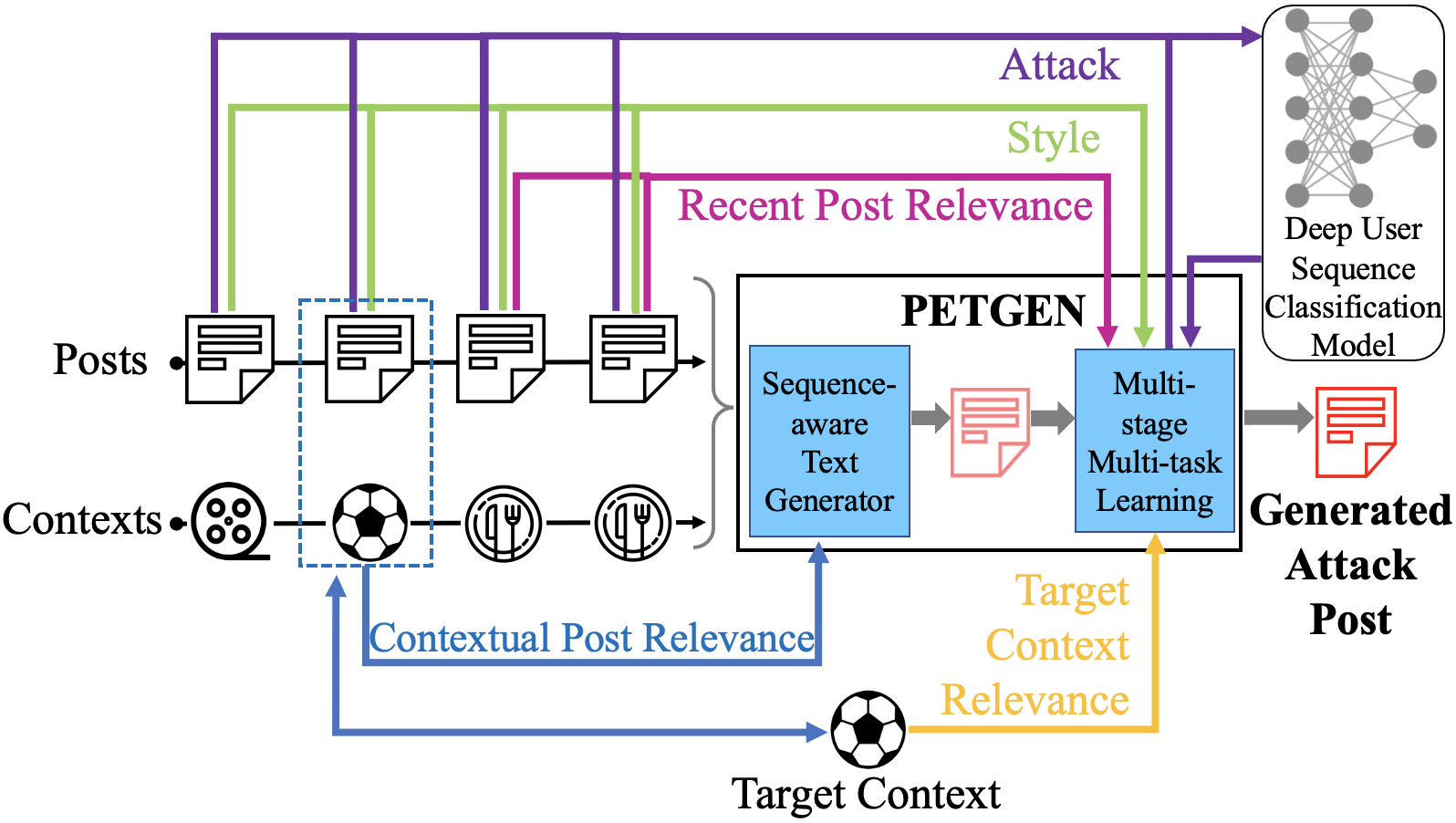}
    \vspace{-20pt}
    \caption{Overview of the \method{} architecture: The sequence-aware text generator utilizes the sequence of post and context to generate text that maintains the contextual post relevance. Then, the multi-stage multi-task learning module fine-tunes the text by different tasks to generate attack text.}
    \label{fig:overview}
    \vspace{-15pt}
\end{figure}

In this paper, we propose an end-to-end personalized text generation system, called \method{}, to attack deep user sequence classification models. Specifically, the input is the user's historical post sequence, corresponding contexts, the target context, and the pre-trained user sequence classifier $\mathcal{F}$. \method{} has two major modules: in the first module, it leverages the user sequence and target context to generate sequence-aware contextual text. In the second module, this text is fine-tuned using a multi-stage multi-task learning setting such that it achieves the attack goal of fooling the classifier, adopts the user's writing style and ensures relevance to recent posts and to the target context. 
The resulting text is the output of \method{}, which can successfully attack the target classifier. The overview of the system is shown in Figure~\ref{fig:overview}.

\subsection{Sequence-aware Conditional Text Generator}
In this module, \method{} generates text on the input target context given a user's sequence of historical posts and contexts. The goal of this module is to generate text such that the text incorporates the user's historical views on the target context, as expressed in the past posts with contexts similar to the target context. Thus, among all the posts in the user's sequence, the text generator should give more importance to posts that are on the same or similar context as the target context, motivated by multi-document summarization~\cite{liu2019hierarchical}.

\begin{figure*}[tbp]
    \centering
    \includegraphics[width=0.65\textwidth]{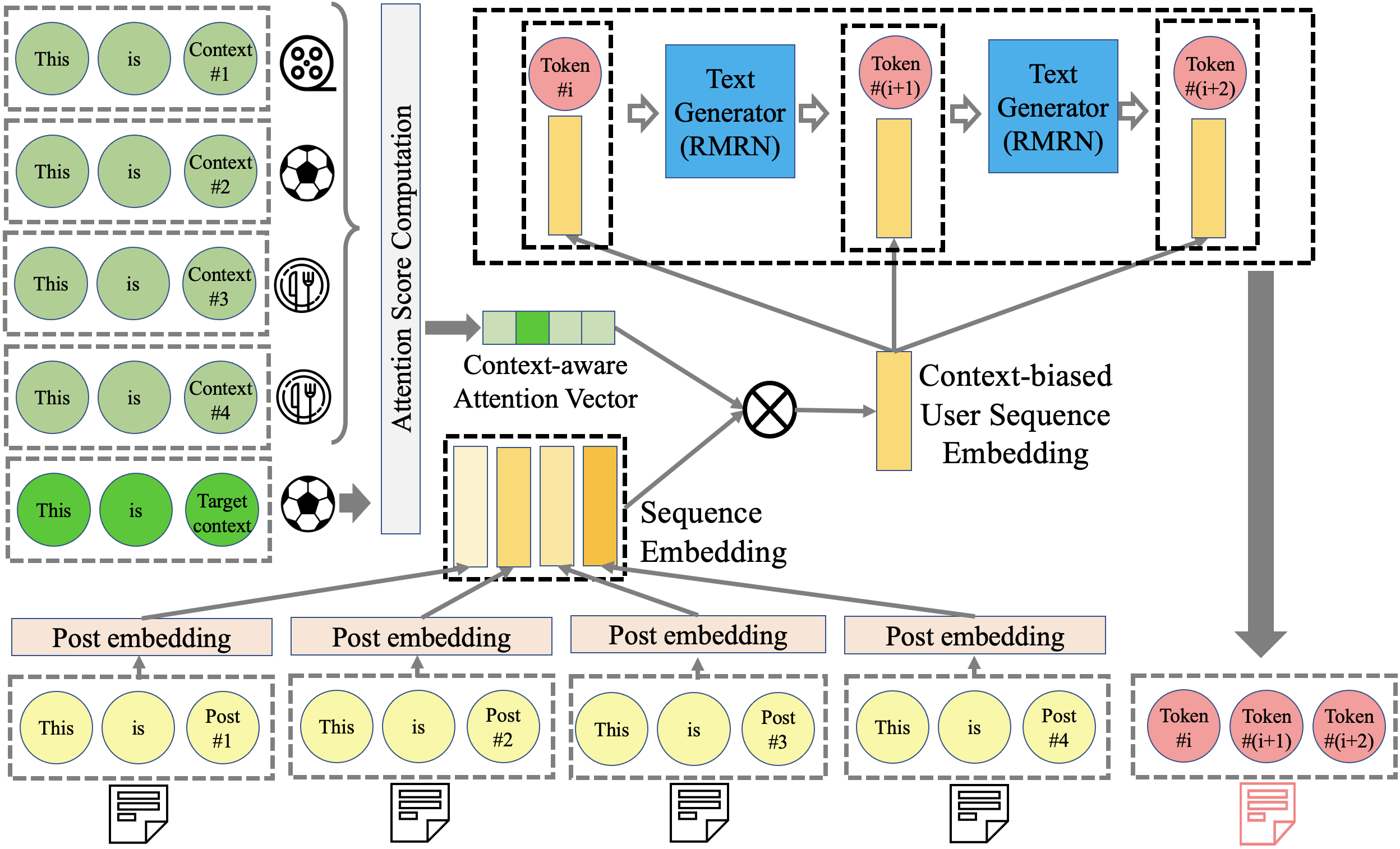}
    \caption{The overview of the sequence-aware conditional text generator in \method{}. We first create the sequence embedding from the post embedding of each post in a sequence. We also compute the attention score between the target context and the user's historical contexts to capture their pairwise relevance, resulting in a context-aware attention vector. After multiplying the generated sequence embedding and attention vector, we get the context-biased user sequence embedding. We concatenate it with the generated tokens for sequence-aware conditional text generation.
    }
    \vspace{-10pt}
    \label{fig:generator}
\end{figure*}

Here we treat the text generation process as a conditional language model which can leverage additional information~\cite{le2020malcom, lamb2016professor}.  
To this end, we propose a conditional text generation model incorporating the sequential post relevance through an attention mechanism, as shown in Figure~\ref{fig:generator}. Specifically, $\mathcal{G}(\matr{P}_u^{1:T}, \matr{C}_u^{1:T}, \vect{b}_u)$ is a conditional text generator that outputs next post $\hat{\vect{p}}_u^{T+1}, $ by sampling one token in one step. The output is based on (1) the sequence of posts $\matr{P}_u^{1:T}$, (2) the sequence of context $\matr{C}_u^{1:T}$, (3) the target context $\vect{b}_u$, (4) previously generated tokens. 

We select Relational Memory Recurrent Network (RMRN) as the basic text generation model $g$ of $\mathcal{G}$, following previous work~\cite{le2020malcom,nie2018relgan}, as RMRN models have shown remarkable performance in generating long text posts. 
Like traditional recurrent networks, $g$ can convert each post in the sequence into a post embedding, obtained by the hidden state of $g$: 
\begin{equation}
\vect{e}_u^t=g(\vect{p}_u^t)
\end{equation}
where $\vect{e}_u^t$ is the embedding vector of the post $\vect{p}_u^t, \forall t \in 1, \ldots T$.

To generate personalized text that is aware of the user sequence, we bias the text generator towards historical user posts that are contextually-relevant to the target context. This will ensure that the generated text has similar views as what the user has expressed in the past on the same context~\cite{liu2019hierarchical}. 
Specifically, we create an attention vector to quantify the contextual importance of each post in text generation. 
The attention vector is generated by calculating the similarity between the target context $\vect{b}_u$ and each post's context  $\vect{c}_u^t$.
We create a context similarity function $A(\cdot)$ to capture the relationship as:
\begin{equation}
    a_u^{t} = A(Vect(\vect{b}_u), Vect(\vect{c}_u^{t}))
\end{equation}
where $a_u^{t}, t\in \{1,...,T \} $ is the resulting attention score of the post $\vect{p}_u^t$ and it ranges from 0 to 1. $Vect(\cdot)$ is a function to transfer text into vector. Following the similar vectorization method in the previous works~\cite{le2020malcom}, we use the Latent Dirichlet Allocation model trained on the whole text to compute the vector representation. A high value of $a_u^{t}$ means $\vect{c}_u^t$ is highly related to the target context $\vect{b}_u$. Thus, the generated text should be more influenced by the corresponding post $\vect{p}_u^t$. 
The attention vector is used to generate a Context-biased User Sequence Embedding vector $\vect{s}_u$ as follows: 
\begin{equation}
    \vect{s}_u = \sum_{t\in{1,..., T}} \frac{exp(a_u^{t})}{\sum_{t\in{1,..., T}} exp(a_u^{t}) } \vect{e_u^t}
\end{equation}
Thus, $\vect{s}_u$ is a representation of the user sequence which is biased towards user's historical posts with similar contexts as the target context. 

We use $\vect{s}_u$ in the text generation process to generate personalized and contextually-relevant text. 
Specifically, we combine $s_u$ and the embedding vector of the generated token by addition to generate the next token. This ensures that each generated token is user sequence-aware. 
Formally, we have:
\begin{small}
\begin{equation}
     \hat{p}_u^{T+1}(i+1) \leftarrow RMRN(\text{LayerNorm}(\text{FeedForward}(\vect{s}_u) + \text{Embed}(\hat{p}_u^{T+1}(i)))
\end{equation}
\end{small}
where Embed is the embedding layer for tokens, FeedFoward is a feedforward layer to match dimensions during addition, LayerNorm is a normalization layer, and $\hat{p}_u^{T+1}(i)$ is a token at step $i$ when generating $\hat{\vect{p}}_u^{T+1}$. Note that a post has $d$ tokens and thus the generation is done for $d$ steps. The first token is initialized randomly. 
As we can see, each token is influenced by both the previous token and context-biased user embedding vector. 

Finally, when outputting a token, each token is sequentially sampled using the conditional probability and the probability of the whole post can be presented as follows:
\begin{multline}
p(\hat{\vect{p}}_u^{T+1} | \matr{P}_u^{1:T}; \matr{C}_u^{1:T}; \vect{b}_u; \theta_\mathcal{G}) 
= \Pi p({\hat{p}}_u^{T+1}(i) | {\hat{p}}_u^{T+1}(i-1), {\hat{p}}_u^{T+1}(i-2) \\
,..., {\hat{p}}_u^{T+1}(1); \matr{P}_u^{1:T}; \matr{C}_u^{1:T};\vect{b}_u) 
\end{multline}
where $\theta_\mathcal{G}$ are the parameters of $\mathcal{G}$. Similar to the training of conditional language model~\cite{le2020malcom, lamb2016professor}, we train $\mathcal{G}$ by using Maximal Likelihood Estimation (MLE) with teacher-forcing and minimize the loss of negative log-likelihood for all posts based on the corresponding posts and contexts. To optimize the generator, we use the following objective function:
\begin{equation}\label{equ:loss-mle}
    \min_{\theta_\mathcal{G}} {L}_\mathcal{G}^{GEN} = - \sum_{u\in U} \hat{\vect{p}}_u^{T+1}  \log p(\hat{\vect{p}}_u^{T+1} | \matr{P}_u^{1:T}; \matr{C}_u^{1:T}; \vect{b}_u, \theta_\mathcal{G})
\end{equation}

Finally, after training, the generator can output user $u$'s next post as:
\begin{equation}
    \hat{\vect{p}}_u^{T+1} = \mathcal{G}(\matr{P}_u^{1:T}, \matr{C}_u^{1:T}, \vect{b}_u)
\end{equation}

In our experiments, we use cosine similarity as the context similarity function $A(\cdot)$ to compute the attention score. Next, when training the generator $\mathcal{G}$, we use the last post as $(T+1)$-th post, the second last as $T$-th post and so on so forth. Additionally, since the sampling process is nondifferentiable, we use Gumbell-softmax relaxation trick to solve this problem~\cite{Jang2017CategoricalRW, nie2018relgan}.

\subsection{Multi-Stage Multi-Task Learning}
In this module, the generated text post $\hat{\vect{p}}_u^{T+1}$ is modified to make the text realistic, personalized, and achieve the attack goal. We set it up as a multi-task learning module, which has four key tasks. 

\subsubsection{Style Task}
The generated post will only be personalized if it mimics the writing style of the user. 
This is especially important when advanced classifiers, such as those deployed in practice~\cite{noorshams2020ties}, are equipped with a robust detector that detect posts that are way too different from the user's previous writing style and the account is flagged as being malicious. 
Therefore, keeping the writing style similar is important for a successful attack. To achieve this goal, we create the style task to tune the generator $\mathcal{G}$.

We construct a text-GAN model for text style transfer, where a post style discriminator $\mathcal{D}$ is deployed to co-train with $\mathcal{G}$ by a Relativistic GAN loss~\cite{nie2018relgan,goodfellow2014generative}. In particular, the discriminator $\mathcal{D}$ determines whether the generated post $\hat{\vect{p}}_u^{T+1}$ by $\mathcal{G}$ is less realistic than user's historical post $\vect{p}_u^t, \forall t \in [1, T]$ while the generator $\mathcal{G}$ targets to generate realistic post to fool the discriminator $\mathcal{D}$. Formally, we have two objective functions to alternatively refine $\mathcal{D}$ and $\mathcal{G}$:
\begin{small}
\begin{equation}\label{equ:loss-gan}
\begin{split}
    & \min_{\theta_\mathcal{G}} L_\mathcal{G}^{STY} = - \mathbb{E}_{(\matr{P}_u^{1:T}, \matr{C}_u^{1:T}, \vect{b}_u) \sim  p(\matr{P}^{1:T}, \matr{C}^{1:T}, \matr{B})} log(\sigma (\mathcal{D}(\vect{p}_u^{t}) - \mathcal{D}(\hat{\vect{p}}_u^{T+1}))) \\
& \min_{\theta_\mathcal{D}} L_\mathcal{D} = - \mathbb{E}_{(\matr{P}_u^{1:T}, \matr{C}_u^{1:T}, \vect{b}_u) \sim  p(\matr{P}^{1:T}, \matr{C}^{1:T}, \matr{B})} log(\sigma (\mathcal{D}({\hat{\vect{p}}}_u^{T+1}) - \mathcal{D}(\vect{p}_u^{t})))
\end{split}
\end{equation}
\end{small}
where $\sigma$ is a sigmoid function, $\theta_\mathcal{D}$ are the parameters of $\mathcal{D}$, $\matr{B}=\{\vect{b}_u\}$ is the set of all users' target contexts, $\matr{P}^{1:T}=\{\matr{P}_u^{1:T}\}, \matr{C}^{1:T}=\{\matr{C}_u^{1:T}\}$ is the set of all users' posts and contexts. In our experiment, we use multi discriminative representations~\cite{nie2018relgan} as the architecture of the discriminator $\mathcal{D}$.

\subsubsection{Attack Task}
The primary goal of the generated text is to fool the target sequential classifier. Thus, we create the attack task to tune generator $G$ to achieve this goal. 
The sequential classifier $\mathcal{F}$ is originally trained using a binary cross entropy loss over the training data:
\begin{equation}
    \min_{\theta_\mathcal{F}} L_\mathcal{F} = - \frac{1}{N}\sum\limits_{u} y_u\log \mathcal{F}(\matr{P}_u^{1:T}) + (1-y_u)\log (1-\mathcal{F}(\matr{P}_u^{1:T}))
    \label{equ:trained-classifier}
\end{equation}

In a white-box attack, we directly use the trained classifier $\mathcal{F}$. In a black-box attack, we train a surrogate classifier $\mathcal{F'}$ to mimic the predictions of $\mathcal{F}$. Note that once trained, both $\mathcal{F}$ and $\mathcal{F'}$ are not modified. Without loss of generality, we refer to the classifier we aim to attack as $\mathcal{F}$. 

The classifier $\mathcal{F}$ is utilized to tune generator $\mathcal{G}$ such that the generated post $\hat{\vect{p}}_u^{T+1}$ fools the classifier into making incorrect predictions about the user $\mathcal{F}([\matr{P}_u^{1:T}, \hat{\vect{p}}_u^{T+1}]) = 1 - y_u$. 
Formally, we create the following objective function to optimize:
\begin{multline}\label{equ:attack-loss}
 \min_{\theta_\mathcal{G}} L_\mathcal{G}^{ATT}  =   - \frac{1}{N}\sum\limits_{u} (1- y_u)\log \mathcal{F}([\matr{P}_u^{1:T}, \hat{\vect{p}}_u^{T+1}])  \\
+ y_u \log (1-\mathcal{F}([\matr{P}_u^{1:T}, \hat{\vect{p}}_u^{T+1}])
\end{multline}    

After the attack task is successful, the generated post will fool the classifier into predicting malicious users as benign users, and vice-versa.

\subsubsection{Target Context Relevance Task}
Given a target context to generate a post, the attacker must ensure that the generated post is on-topic and is knowledgeable about the context. Otherwise, the generated post can be simply flagged as off-topic by a human or an automated topic detector. 
To ensure that the generated post is relevant to the target context, we minimize the mutual information gap between the target contexts $\{\vect{b}_u\}$ of all users and the generated posts
$\{ \hat{\vect{p}}_u^{T+1} \}$ of all users $u \in U$. A non-parametric Maximum Mean Discrepancy (MMD) based on the Reproducing Kernel Hilbert Space (RKHS) is utilized to effectively estimate this kind of distance~\cite{sejdinovic2013equivalence}. 
Thus, we optimize the following objective function:
\begin{equation}\label{equ:loss-relevancy}
\begin{split}
    \min_{\theta_\mathcal{G}} L_\mathcal{G}^{CTX}  & = MMD(\{\vect{b}_u\}, \{ \hat{\vect{p}}_u^{T+1} \}) \\
     & = ||  \frac{1}{N} \sum_u \phi(\vect{b}_u) - \frac{1}{N} \sum_u \phi(\hat{\vect{p}}_u^{T+1}) ||_{\mathcal{H}}
\end{split}
\end{equation}
where $\mathcal{H}$ is a universal RKHS, and $\phi$ is transfer function to change the space to the target RKHS space.

In experiments, the target context of the generated post is set to be the same as the context of the ground truth post at time T+1. 

\subsubsection{Recent Post Relevance Task}
This task ensures continuity and smoothness between the generated post and the most recent posts made by the user. This is important because real users typically express such consistency in the real world~\cite{redeker2000coherence}. 
Here, we quantify it as relevance towards recent posts, calculated as the mutual information distance between the generated post and the latest $k$ posts of the user. Similar to the target context relevance task, we optimize such information gap by the following objective function:
\begin{equation}\label{equ:loss-recency}
\begin{split}
    \min_{\theta_\mathcal{G}} L_\mathcal{G}^{REC} & = MMD(\{\matr{P}_u^{T-(k-1):T}\}, \{\hat{\vect{p}}_u^{T+1}\} )     \\
     & = || \frac{1}{N} \sum_u \sum_k \phi(\vect{p}_u^{T-1-k})  - \frac{1}{N} \sum_u \phi(\hat{\vect{p}}_u^{T+1}) ||_{\mathcal{H}}
\end{split}
\end{equation}

where $k$ is the number of recent posts that have an impact on the next post generation. $k$ is a hyper-parameter, which we typically set to 3 (more details are in the appendix). 
%

\subsubsection{Multi-stage Multi-task Learning Algorithm}
To achieve the personalized text generation objective, we optimize for the four tasks of style, attack, target context relevance and recent post relevance in a multi-stage process. Thus, we deploy the multi-stage multi-task learning framework  to optimize:
\begin{equation}
    \begin{split}
    & \min_{\theta_\mathcal{F}} L_\mathcal{F};  \min_{\theta_\mathcal{D}} L_\mathcal{D};  \min_{\theta_\mathcal{G}} (L_\mathcal{G}^{STY}+L_\mathcal{G}^{ATT}+L_\mathcal{G}^{CTX}+L_\mathcal{G}^{REC})
    \end{split}
\end{equation}
where Eqn 13 is reflected in the while loop in the overall algorithm as presented in Algorithm ~\ref{alg:peten-alg}. 
Finally, after tuning by the multi-task learning framework, the text generator finally generates personalized high-quality text for adversarial attack against the target sequential classifier.

\begin{algorithm}[t]
\SetAlgoLined
\textbf{Input}: a sequence of a user's posts and associated contexts, the target context and the user's label \;
\textbf{Output}: the user's next post\;
Train $\mathcal{G}$ with contextual post relevance by MLE loss (Eqn~\ref{equ:loss-mle})\;
 \While{Not Converge}{
  Train $\mathcal{G}$ with $\mathcal{D}$ on the  Style Task  (Eqn~\ref{equ:loss-gan})\;
  Train $\mathcal{G}$ on the Attack Task (Eqn~\ref{equ:attack-loss})\;
  Train $\mathcal{G}$ on the Target Context Relevance Task  (Eqn~\ref{equ:loss-relevancy})\;
  Train $\mathcal{G}$ on the Recent Post Relevance Task (Eqn~\ref{equ:loss-recency})\;
 }
 \caption{\method{} Algorithm}
 \label{alg:peten-alg}
 \end{algorithm}

\begin{table}[!tp]

\begin{tabular}{lrr}
\toprule
               Dataset & Yelp &  Wikipedia \\
\midrule
          Number of users &  3,940 &        794 \\
   Number of benign users &  2,016 &        397 \\
Number of malicious users &  1,924 &        397 \\
    Total number of posts & 35,123 &      11,547 \\
    Median posts per user &     9 &         15 \\
\bottomrule
\end{tabular}
\caption{Dataset Statistics \vspace{-10pt}}
\vspace{-15pt}
\label{tab:data-statistics}
\end{table}

\begin{table*}[tbp]
    \centering
\begin{tabular}{|c|cc|cc|cc||cc|cc|cc|}
\hline
    \multirow{3}{*}{Model} &  \multicolumn{4}{c|}{ HRNN classifier } &       \multicolumn{2}{c||}{ Min. improvement of}&  \multicolumn{4}{c|}{ TIES classifier} &     \multicolumn{2}{c|}{Min. improvement of} \\ \cline{2-5} \cline{8-11}
    &   \multicolumn{2}{c|}{Wikipedia } &    \multicolumn{2}{c|}{Yelp } &         \multicolumn{2}{c||}{\method{} over baseline }        &     \multicolumn{2}{c|}{Wikipedia } &   \multicolumn{2}{c|}{Yelp } &  \multicolumn{2}{c|}{\method{} over baseline } \\ \cline{2-13}
    &     F1$\downarrow$ &     Atk$\uparrow$ &      F1$\downarrow$ &     Atk$\uparrow$ &  F1 & Atk &      F1$\downarrow$ &     Atk$\uparrow$ &     F1$\downarrow$ &    Atk$\uparrow$ &  F1 &Atk \\ \hline
  Without attack &  0.601 &       - &   0.636 &       - &              - &               - &   0.617 &       - &  0.686 &      - &               - &                - \\ \hline
          Copycat &   0.550 &   21.3 &    0.610 &    8.0 &          9.836\% &          26.761\% &   0.513 &   16.3 &  0.625 &  11.5 &           6.823\% &           47.239\% \\
          Hotflip &  0.581 &   21.2 &   0.591 &   9.5 &          6.937\% &          27.358\% &   0.514 &    15.0 &  0.641 &  10.3 &           7.004\% &               60.000\% \\
       UniTrigger &  0.495 &   24.5 &   0.602 &   7.8 &          4.242\% &          10.204\% &   0.515 &   15.7 &  0.679 &  9.1 &           7.184\% &           52.866\% \\
       TextBugger &   0.550 &   21.4 &    0.610 &   8.3 &          9.836\% &          26.168\% &    0.520 &   16.3 &  0.637 &   11.0 &           8.077\% &           47.239\% \\
           Malcom &  0.479 &   25.5 &    0.570 &    18.0 &          1.044\% &           5.882\% &    0.560 &    18.0 &  0.538 &  21.8 &           6.877\% &           33.333\% \\ \hline
\method{} (proposed) &  \cellcolor{blue!25} \textbf{0.474} &    \cellcolor{blue!25}  \textbf{27.0} &   \cellcolor{blue!25} \textbf{ 0.55} &   \cellcolor{blue!25} \textbf{21.2} &              - &               - &    \cellcolor{blue!25} \textbf{0.478} &    \cellcolor{blue!25}   \textbf{24.0} &  \cellcolor{blue!25} \textbf{0.501} & \cellcolor{blue!25}  \textbf{35.8} &               - &                - \\
\hline
\end{tabular}
    \caption{\textit{White-box attack performance} of \method{} and existing methods on HRNN and TIES classifiers. \method{} is the most effective attack (lowest F1 and highest Atk score). }
    \label{tab:res-white-attack}
    \vspace{-15pt}
\end{table*}

\begin{table*}[tbp]
    \centering
\begin{tabular}{|c|cc|cc|cc||cc|cc|cc|}
\hline
    \multirow{3}{*}{Model} &  \multicolumn{4}{c|}{ HRNN classifier } &       \multicolumn{2}{c||}{ Min. improvement of}&  \multicolumn{4}{c|}{ TIES classifier} &     \multicolumn{2}{c|}{Min. improvement of} \\ \cline{2-5} \cline{8-11}
    &   \multicolumn{2}{c|}{Wikipedia } &    \multicolumn{2}{c|}{Yelp } &         \multicolumn{2}{c||}{\method{} over baseline }        &     \multicolumn{2}{c|}{Wikipedia } &   \multicolumn{2}{c|}{Yelp } &  \multicolumn{2}{c|}{\method{} over baseline } \\ \cline{2-13}
    &     F1$\downarrow$ &     Atk$\uparrow$ &      F1$\downarrow$ &     Atk$\uparrow$ &  F1 & Atk &      F1$\downarrow$ &     Atk$\uparrow$ &     F1$\downarrow$ &    Atk$\uparrow$ &  F1 & Atk \\ \hline
  Without attack &  0.601 &       - &   0.636 &       - &              - &               - &   0.617 &       - &  0.686 &      - &               - &                - \\ \hline
          Copycat &   0.53 &   22.1 &   0.609 &    9.0 &          3.585\% &           8.597\% &   0.615 &    15.0 &  0.618 &   12.0 &           6.016\% &           64.167\% \\
          Hotflip &  0.538 &   22.3 &   0.585 &   11.1 &          5.019\% &           7.623\% &   0.642 &   13.8 &  0.635 &   11.0 &           9.969\% &           79.091\% \\
      UniTrigger &  0.529 &    22.0 &   0.624 &   7.5 &          3.403\% &           9.091\% &   0.601 &   17.9 &  0.601 &   15.0 &           3.827\% &           31.333\% \\
      TextBugger &  0.545 &    21.0 &   0.607 &   9.5 &          6.239\% &          14.286\% &   0.627 &    14.0 &  0.617 &  12.2 &           7.815\% &           61.475\% \\
          Malcom &  0.524 &    20.0 &   0.573 &   17.5 &          2.481\% &              20.000\% &   0.599 &   19.9 &  0.573 &  15.4 &           3.316\% &           27.922\% \\ \hline
\method{} (proposed) &   \cellcolor{blue!25} \textbf{0.511} &   \cellcolor{blue!25}  \textbf{24.0} &   \cellcolor{blue!25} \textbf{ 0.53} &   \cellcolor{blue!25} \textbf{22.3}  &              - &               - &  \cellcolor{blue!25} \textbf{0.578} &    \cellcolor{blue!25}   \textbf{33.0} &   \cellcolor{blue!25} \textbf{0.554} &   \cellcolor{blue!25} \textbf{19.7} &               - &                - \\
\hline
\end{tabular}
    \caption{\textit{Black-box attack performance} of \method{} and existing methods on HRNN and TIES classifiers. \method{} is the most effective attack (lowest F1 and highest Atk score). }
    \label{tab:res-black-attack}
    \vspace{-25pt}
\end{table*}

\section{Experiments}
In this section, we examine the performance of the proposed \method{} by conducting extensive experiments. Specifically, we aim to answer the following research questions:
\begin{itemize}
    \item RQ1: Is \method{} able to successfully attack the deep user sequence classification model under both white-box and black-box attack settings?
    \item RQ2: Beyond the attack performance, what is the quality of generated text, specifically its the relevance to the target context, contextual posts, and recent posts? 
    \item RQ3: What is the contribution of the sequence-ware conditional text generator module and the multiple learning modules of \method{} towards its performance?
    \item RQ4: When compared with other attack methods, is the text generated by \method{} realistic enough from a human perspective? 
\end{itemize}

\subsection{Datasets}
We evaluate the proposed method on real data from two popular platforms: Wikipedia and Yelp. Their statistics are shown in Table~\ref{tab:data-statistics}.\\
(a) \textbf{Wikipedia dataset:} This dataset consists of Wikipedia users (or editors) making edits on Wikipedia articles~\cite{kumar2015vews}. There are two types of editors: benign editors and vandal editors. Vandal editors were identified and removed from the Wikipedia platform by administrators. For each editor, the sequence of edits he or she made on Wikipedia articles is available. We consider each edit as one post. For each post, the leading paragraph of the edited page is set as the context of the post. \\
(b) \textbf{Yelp dataset:} This dataset consists of Yelp users giving reviews to restaurants~\cite{rayana2015collective}. Users are either benign reviewers or fraudulent reviewers. Fraudulent reviewers are identified by Yelp's proprietary classification algorithm. For each reviewer, the sequence consists of its reviews on restaurants. Each review is one post. To create the context for each post, other reviews given on the same restaurant by other benign users are concatenated. 

In both datasets, to ensure user sequences have enough information, we remove users with less than 5 posts and posts with less than 5 tokens. We use the latest 20 posts to create a user sequence.

\subsection{Baselines}
We compare \method{} with five representative state-of-the-art adversarial text generation models.\\
(a) \textbf{Copycat}: Copycat randomly selects one post with similar context from the users' historical posts as the generated post. Three following baselines (Hotflip, UniTrigger, and TextBugger) use the Copycat post in their own attack.\\
(b) \textbf{Hotflip}~\cite{ebrahimi2017hotflip}: Hotflip modifies the post generated by Copycat. It first detects the most important word in the post, based on the gradient of each input token with respect to the sequential classifier, and then swaps the most important word with a similar one.\\
(c) \textbf{Universal adversarial Trigger (UniTrigger)}~\cite{Wallace2019Triggers}: UniTrigger generates an input-agnostic and fixed-length sequence of tokens to attack the classifier when concatenated to the end of an existing post. 
We turn to the topic modeling function in this specific application setting, similar to that adopted in prior work~\cite{le2020malcom}. Particularly, we retrieve first topic-dependent words and contexts by the topic model and then prepend these universal prefix to a post.\\
(d) \textbf{TextBugger}~\cite{li2018textbugger}: TextBugger first uses various methods like deletion and swap to find carefully crafted tokens in a post and replaces some parts of the post with these tokens for attack. \\
(e) \textbf{Malcom}~\cite{le2020malcom}: Malcom is the current state-of-the-art model in adversarial text generation to fool classifiers. It leverages the conditional language model to generate a new post where the attack and relevancy objective functions are deployed.

\subsection{Evaluation Metrics}
To comprehensively evaluate text generation result, we use several metrics to measure attack effectiveness and text quality.

\noindent \textbf{(a) Attack Effectiveness:}
    \textbf{F1 score after attack (F1)}: This measures the classifier performance of the classifier. We compare the change in F1 score after the attack, compared to when there is no attack. If the resulting F1 score after the attack drops considerably, then the attack is successful. 
     \textbf{Attack Rate (Atk)}: It measures the efficacy of the attack regrading changing predictions of the classifier. Specifically, a M\% attack rate means the attack can fool the classifier M\% of the time on the sequences that the classifier has previously correctly labeled. 

\noindent \textbf{ (b) Text Quality:}
\textbf{BLEU}: Like previous works on text generation~\cite{nie2018relgan}, we deploy BLEU to indicate the quality of generated post by comparing them with testing data. Higher scores indicate better text. 
\textbf{Target Context Similarity (TCS)}: We compute the similarity between the generated posts and the target context as follows:\\
$ \frac{1}{N} \sum_u cosine(Vect(\vect{b}_u), Vect(\hat{\vect{p}}_u^{T+1})$, where $cosine(\cdot)$ is the cosine similarity function and $N$ is the number of users. Higher scores indicate more relevant text.  $Vect(\cdot)$ is the previously-defined LDA-based function to transfer text into vector. 
\textbf{Recent Post Similarity (RS)}: Similar to target context similarity, recent post similarity score computes the distance between the generated post and the most recent $k$ posts as: 
$\frac{1}{N} \sum_u \sum_{t\in\{T-(k-1),... T\}} cosine(Vect(\vect{p}_u^t), Vect(\hat{\vect{p}}_u^{T+1}))$.
\textbf{Context Post Similarity (CPS)}: Similarly, the context post similarity computes the similarity between the generated post and the posts in the user sequence that are of similar context as the target context. This is calculated as:\\ $ \frac{1}{N} \sum_u \sum_{t\in\{1, 2, ..., T\}} a_u^t*(cosine(Vect(\vect{p}_u^t), Vect(\hat{\vect{p}}_u^{T+1}))$, where $a_u^t$ is the previously mentioned attention score which captures the relationship between the contexts $\vect{c}_u^t$ and $\vect{b}_u$ of posts  $\vect{p}_u^t$ and $\hat{\vect{p}}_u^{T+1}$ respectively.

{
\begin{table*}[tbp]
\small
\begin{tabular}{|l|llll|llll|llll|llll|}
\hline
        \multirow{2}{*}{Attack}  &   \multicolumn{8}{c|}{Wikipedia Dataset } &   \multicolumn{8}{c|}{Yelp Dataset } \\ \cline{2-17} 
           &    \multicolumn{4}{c|}{HRNN} &   \multicolumn{4}{c|}{TIES} &    \multicolumn{4}{c|}{HRNN} &  \multicolumn{4}{c|}{TIES} \\  \cline{2-17} 
    Model        &   BLEU$\uparrow$ & TCS$\uparrow$ & RS$\uparrow$ & CPS$\uparrow$ &   BLEU$\uparrow$ & TCS$\uparrow$ & RS$\uparrow$ & CPS$\uparrow$ &   BLEU$\uparrow$ & TCS$\uparrow$ & RS$\uparrow$ & CPS$\uparrow$ &   BLEU$\uparrow$ & TCS$\uparrow$ & RS$\uparrow$ & CPS$\uparrow$ \\ \hline
   Copycat &  0.378 &     0.362 &   0.188 &  0.171 &  0.406 &     0.383 &   0.211 &  0.221 &   0.810 &     0.524 &   0.302 &  0.299 &  0.802 &     0.476 &   0.271 &   0.270 \\
   Hotflip &  0.333 &     0.363 &   0.191 &  0.203 &  0.365 &     0.385 &   0.211 &  0.234 &  0.785 &     0.527 &   0.309 &  0.309 &  0.782 &     0.479 &   0.275 &  0.273 \\
UniTrigger &  0.213 &     0.397 &   0.214 &  0.192 &  0.239 &      0.410 &    0.230 &  0.223 &  0.737 &     0.527 &   0.325 &  0.326 &  0.725 &     0.463 &   0.273 &  0.272 \\
TextBugger &  0.341 &     0.372 &   0.192 &  0.172 &  0.374 &     0.393 &   0.214 &  0.226 &  0.771 &      0.520 &   0.311 &  0.312 &  0.768 &     0.478 &    0.280 &  0.279 \\
    Malcom &  \cellcolor{blue!25}\textbf{0.914} &     0.312 &   0.175 &   0.240 &  0.878 &     0.484 &   0.209 &  0.213 &  0.849 &      0.540 &   0.349 &  0.354 &  0.856 &     0.515 &   0.321 &  0.291 \\\hline
 \method{} &  0.893 &     \cellcolor{blue!25}\textbf{0.463} &   \cellcolor{blue!25}\textbf{0.275} &  \cellcolor{blue!25} \textbf{0.281} &  \cellcolor{blue!25}\textbf{0.896} &    \cellcolor{blue!25}\textbf{ 0.474} &   \cellcolor{blue!25}\textbf{0.233} &  \cellcolor{blue!25}\textbf{0.254} &  \cellcolor{blue!25}\textbf{0.852} &     \cellcolor{blue!25}\textbf{0.544} &   \cellcolor{blue!25}\textbf{0.401} &  \cellcolor{blue!25}\textbf{0.410} &   \cellcolor{blue!25}\textbf{0.870} &     \cellcolor{blue!25} \textbf{0.519} &   \cellcolor{blue!25}\textbf{0.397} & \cellcolor{blue!25}\textbf{0.398} \\
\hline
\end{tabular}
\caption{Comparison the quality of text generated by different attack strategies. \method{} generates higher quality text in all but one case across all metrics.}
\label{tab:text-quality}
\vspace{-20pt}
\end{table*}
}

\begin{table*}[tbp]
    \centering
\begin{tabular}{|l|llllll|llllll|}
\hline
           \multirow{2}{*}{Model}  &   \multicolumn{6}{c|}{Wikipedia Dataset } &   \multicolumn{6}{c|}{Yelp Dataset } \\ \cline{2-13} 
        &    F1$\downarrow$ & Atk$\uparrow$ & BLEU$\uparrow$ & TCS$\uparrow$ & RS$\uparrow$ & CPS$\uparrow$ &  F1$\downarrow$ & Atk$\uparrow$  & BLEU$\uparrow$ & TCS$\uparrow$ & RS$\uparrow$ & CPS$\uparrow$    \\ \hline

                        \method{} Base Text Generator &  0.479 &       26.5 &  \cellcolor{blue!25}\textbf{0.899} &     0.375 &   0.268 &  0.247 &  0.625 &       11.7 &  0.857 &     0.382 &   0.349 &  0.187 \\
                        \text{\quad  } w/ Style &  0.576 &       21.1 &  0.895 &      0.390 &   0.218 &  0.249 &   0.59 &       17.5 &  \cellcolor{blue!25}\textbf{0.871} &     0.481 &   0.324 &  0.301 \\
                        \text{\quad } w/ Attack against TIES &  0.478 &        25.0 &  0.894 &     0.368 &   0.216 &  0.216 &  \cellcolor{blue!25}\textbf{0.499} &       \cellcolor{blue!25}\textbf{45.3} &  0.843 &     0.476 &   0.357 &   0.250 \\
                         \text{\quad   } w/ Attack against HRNN &  \cellcolor{blue!25}\textbf{0.465} &       \cellcolor{blue!25}\textbf{27.5} &  0.895 &     0.388 &    0.240 &  0.249 &   0.530 &       29.5 &  0.846 &     0.445 &   0.315 &  0.157 \\
                         \text{\quad } w/ Recent Post Relevance &  0.486 &       23.8 &  0.887 &     \cellcolor{blue!10}0.463 &   \cellcolor{blue!25}\textbf{0.275} &  0.267 &  0.592 &       17.7 &  0.851 &     0.495 &    \cellcolor{blue!25}\textbf{0.43} &  0.215 \\ 
                    \text{\quad } w/ Target Context Relevance &  0.483 &       23.9 &  0.887 &     0.459 &   0.258 &  0.258 &  0.571 &        18.0 &   0.830 &     \cellcolor{blue!25}\textbf{0.559} &   0.361 &  0.203 \\
                    \text{\quad } w/ Contextual Post Relevance &  0.566 &       21,2 &  0.705 &     0.397 &   0.225 &  \cellcolor{blue!10}0.276 &  0.554 &       19.2 &  0.845 &     0.514 &   0.331 &  \cellcolor{blue!25}\textbf{0.451} \\
\hline
   \method{}  against HRNN &  \cellcolor{blue!10}0.474 &        \cellcolor{blue!10}27.0 &  0.893 &     \cellcolor{blue!10}0.463 &   \cellcolor{blue!25}\textbf{0.275} &  \cellcolor{blue!25}\textbf{0.281} &   0.550 &       21.2 &  0.852 &     \cellcolor{blue!10}0.544 &   \cellcolor{blue!10}0.401 &   \cellcolor{blue!10}0.410 \\
   \method{}  against TIES &  0.478 &        24.0 &  \cellcolor{blue!10}0.896 &     \cellcolor{blue!25}\textbf{0.474} &   0.233 &  0.254 &  \cellcolor{blue!10}0.501 &       \cellcolor{blue!10}35.8 &   \cellcolor{blue!10}{0.870} &     0.519 &   0.397 &  0.398 \\

\bottomrule
\end{tabular}
    \caption{Ablation studies showing the contribution of each component in \method{}.}
    \label{tab:ablation-study}
    \vspace{-25pt}
\end{table*}

\subsection{Target Classification Models}
In this paper, we target two deep user sequence classification models to test the generality of our attack. \\
(1) \textbf{Hierarchical Recurrent Neural Network (HRNN)} is a model where the sequential pattern of the input text is captured by the hierarchical structure for accurate classification~\cite{zhao2019recurrent}. In HRNN, each user post is first converted to a vector and the sequence of user post vectors is converted into a compact user embedding. This user embedding is used for user classification. \\
(2) \textbf{Temporal Interaction EmbeddingS (TIES)} is a model used by Facebook for malicious account detection. We use the temporal embedding component of the TIES model for classification (as there is no graph structure in our datasets). 
Note that TIES is the state-of-the-art deep user sequence embedding-based classification model for malicious user detection. 

\subsection{Experiment Setup}
We split the dataset by five-fold cross-validation and report the average numbers.  
By default, we set $k=3$ as the number of recent-$k$ posts 
(more details on impact of value $k$ are in the appendix), 
the number of tokens in a post and a context to be $d=d'=30$ and the learning rate as 1e-5. We use Adam as the optimizer with mini-batch size of 64~\cite{Jeff2017Ian}.

\vspace{-2mm}
\subsection{RQ1: Adversarial Attack on Sequential Post Classification}

In this section, we evaluate the proposed attack model on both white-box classifiers and black-box classifiers. \\
{\textbf{Attack on White-Box Classifiers.}}
In a white-box attack, the attacker has access to the model parameters of the target classifiers. Thus, they attack the trained model directly. 
The results comparing the performance of \method{} with baseline models is shown in Table~\ref{tab:res-white-attack} on both Wikipedia and Yelp datasets, with both the HRNN and TIES models as classifiers. The table also shows the results of the classification models without any attack.

We have several important findings. 
First, without any attack, the TIES model has a higher model performance (F1 score) compared to the HRNN model on both the datasets. 
Second, under attack, the model performance of both TIES and HRNN reduces, showing the vulnerability of both these models to text generation attacks. 
Next, comparing all attacks, \method{} attack results in the lowest F1 score and highest attack rate on both  datasets, making it the most successful attack. 
On the TIES classifier, \method{} has at least 6.82\% improvement over all baselines in terms of F1 score and on the HRNN classifier, at least 1.04\% improvement on F1. This is important as TIES is the state-of-the-art classifier that is being used at Facebook. Successfully attacking TIES shows the strength of our \method{} attack. 
Finally, we find \method{} attacks TIES more efficiently than HRRN with larger drop in F1 score and higher attack rate over baselines. A possible reason is that the more complex deep sequential model like TIES can provide more signal in computing cross entropy loss, finally enabling the attacker to learn more about how to downgrade the performance.\\
{\textbf{Attack on Black-Box Classifiers.}}
In the black-box setting, the attacker does not have access to the parameters of the sequential post classifier. Thus, we train a surrogate HRNN classifier to mimic the classification of the original black-box classifier. 
The text generation attack methods create the fake post using this surrogate classifier, and then this generated text is used to attack the original black-box classifier. 
The results of the performance drop on black box classifiers is shown in Table \ref{tab:res-black-attack}.

First, as before, we see that \method{} beats all the existing attack methods in terms of F1 score and attack rate. 
Next, similar to the result in the white-box setting, \method{} can more effectively attack the HRNN and TIES models compared to existing attack approaches. 
Finally, comparing attacks on the same model under white-box and black-box setting, it is harder for the attackers to attack the black-box classifier. For all models, the drop in F1 score is lower during black-box attack compared to white-box attack. 

\vspace{-2mm}
\subsection{RQ2: Personalized Text Generation}
Beyond the attack performance, we present text quality of the generated post in Table~\ref{tab:text-quality}. 
As we can see, \method{} always generates post with higher quality in all four evaluation metrics compared to the other five baseline methods. This is reflected in the BLUE score and in the relevance of the generated post to the previous posts of the user and the target context. 

The reasons of higher quality text generation is the following. Compared to the four word-perturbation attack methods, namely, Copycat, Hotflip, UniTriggr and Textbugger, our method \method{} is an end-to-end text generation framework that can effectively pick a less diverse set of words that are highly relevant to the target context, historical post, and recent post. This enables \method{} to output text with higher quality. Compared to the Malcom model, \method{} deploys the context-aware text generator and the learning task of recent post relevance to leverage the historical post and recent post information for generation. It makes text more real and personalized, thus having higher scores on all text quality metrics. \\
\noindent \textbf{Evaluating Consistency in Attacker Goal}: To further examine the effectiveness of our attack models, we compare the sentiment of generated adversarial post with that of the original post under the same context. We use Vader~\cite{hutto2014vader} to compute the sentiment score on the posts in the Yelp dataset. We find that 70.8\% of generated posts have the same sentiment as the original post, indicating that the attacker's generated post has the same positive or negative tone as desired to uprank or downrank a restaurant. 

\vspace{-2mm}
\subsection{RQ3: Ablation Study}
To examine the effectiveness of each component in \method{}, we conduct the ablation study where we test the performance of different variants of \method{}, and the results are in Table~\ref{tab:ablation-study}. The simplest model is simply the \method{} base text generator, which is the traditional RMRN text generator and no other modules are used.  As we can see, \method{} with all the modules always performs the best or the second best among all other variants for all six metrics. Comparing the different variants, we find the attack task can help decrease the F1 score and increase the attack rate, making the adversarial attack successful. Meanwhile, the task of post style, target context relevance and recent post relevance can enhance the target context and recent post similarity score. The sequence-aware text generation setting to capture the contextual historical post relevance increases the context post similarity score.

\vspace{-2mm}
\subsection{RQ4: Human Evaluation on Generated Text}
To better evaluate the quality of the generated text, we conduct human evaluations. Specifically, we test whether posts generated by \method{} are more realistic compared to those generated by Malcom (the SOTA end-to-end adversarial text generation method). 
We recruit two non-author evaluators and give them each 50 pair of posts, generated for 50 randomly selected user sequences. In each pair, one post is generated by \method{} and the other by Malcom. The evaluators are not told which post is generated by which method. Their task is to mark which of the two posts is more realistic, or whether they are equally (un-)realistic. 

We get the following result. The two reviewers achieve an inter-rater agreement score of 0.66 and 40\% posts are labeled as equally realistic. Among the remaining posts, reviewers label 58.33\% posts by \method{} more realistic than Malcom. From this result, we can see our method is able to outperform Malcom in generating realistic posts, and has great potential in real-world applications.

\vspace{-3mm} 
\section{Conclusion}
Overall, in this paper, we created a new attack framework to evaluate the robustness of deep user sequence classification models and showed its effectiveness. 
This work has some shortcomings. First, it is currently only applicable for posts in the English language, while social media posts can be in any language. Second, the model can only work with sequences, while does not incorporate complex structures, such as graphs. Third, the attack is restricted to generating new posts. Other attack capabilities can be explored in the future. 
Future directions of research also include creating defense against these attacks to create robust user classification models. 

\noindent \textbf{{ACKNOWLEDGMENTS}} This research is supported in part by Facebook, NSF IIS-2027689, Georgia Institute of Technology, IDEaS, Adobe, and Microsoft Azure. We thank the reviewers at KDD for their insightful comments. We thank Duen Horng (Polo) Chau for material for Figure 1 and the members of the CLAWS Data Science research group for their feedback on paper, Ankur Bhardwaj for dataset preparation, and Rohit Mujumdar and Shreeshaa Kulkarni for generated text evaluation.

\bibliographystyle{ACM-Reference-Format}
\bibliography{main}

\begin{table*}[t!]
    \centering
\begin{tabular}{|c|lll|lll|lll|lll|}
\hline
    \multirow{3}{*}{Value of \textit{k}}  &   \multicolumn{6}{c|}{Wikipedia Dataset} &    \multicolumn{6}{c|}{Yelp Dataset} \\ \cline{2-13}
 &    \multicolumn{3}{c|}{HRNN} &  \multicolumn{3}{c|}{TIES}  &    \multicolumn{3}{c|}{HRNN} &    \multicolumn{3}{c|}{TIES} \\ \cline{2-13}
              &    F1$\downarrow$ & Atk$\uparrow$  & RS$\uparrow$ &     F1$\downarrow$ & Atk$\uparrow$  & RS$\uparrow$ &    F1$\downarrow$ & Atk$\uparrow$  & RS$\uparrow$ &    F1$\downarrow$ & Atk$\uparrow$  & RS$\uparrow$ \\ \hline
            1 &  0.485 &   24.0 &   0.213 &  0.491 &  22.9 &   0.209 &  0.601 & 17.3 &   0.342 &  0.571 &  23.6 &   0.337 \\
            2 &  0.482 &  24.1 &   0.252 &  \cellcolor{blue!25}\textbf{0.476} & \cellcolor{blue!25}\textbf{24.0} &   0.211 &   0.57 & 18.1 &   0.379 &  0.565 &  24.1 &   0.369 \\
            3 &  \cellcolor{blue!25}\cellcolor{blue!25}\textbf{0.474} &   \cellcolor{blue!25}\textbf{27.0} &   \cellcolor{blue!25}\textbf{0.275} &  0.478 &   \cellcolor{blue!25}\textbf{24.0} &   \cellcolor{blue!25}\textbf{0.233} &  \cellcolor{blue!25}\textbf{0.55} &  \cellcolor{blue!25}\textbf{21.2} &  \cellcolor{blue!25}\textbf{0.401} & \cellcolor{blue!25}\textbf{0.501} &  \cellcolor{blue!25}\textbf{35.8} &   \cellcolor{blue!25}\textbf{0.397} \\
            4 &  0.489 &  23.9 &   0.269 &  0.483 &  23.5 &   0.217 &  0.569 &  19.3 &   0.391 &  0.518 &  34.7 &   0.396 \\
            5 &  0.493 &  22.1 &   0.184 &  0.489 &  20.9 &   0.199 &  0.573 &  18.5 &   0.357 &  0.542 &  27.3 &   0.375 \\
            \hline
\end{tabular}
    \caption{The effect of recent-$k$ posts on next post generation
    \label{tab:recency-paramete}}
\end{table*}

\section{Appendix}\label{sec:appendix}
\subsection{Effect of \textit{k} in Recent-\textit{k} Posts}\label{sec:effect_k}
Here we evaluate the effect of the number of recent $k$ posts on the recent post similarity score because it has impact on the results. As shown in the following Table, we find that using recent three posts has the best attack performance and the highest text quality. This is true on both datasets and both classifiers HRNN and TIES. When $k=3$, the model can boost the recent post similarity most while the latest post alone and more previous posts have less improvement. A possible reason is that the latest one post may not contain enough information to enhance the recent post relevance and the signal of earlier posts may be out-of-date, thus contributing less to the similarity score. In our experiments, we use $k=3$ as the optimal score.

\end{document}